\documentclass[conference]{IEEEtran}
\IEEEoverridecommandlockouts
\usepackage{cite}
\usepackage{amsmath,amssymb,amsfonts}
\usepackage{algorithmic}
\usepackage{graphicx}
\usepackage{textcomp}
\usepackage{xcolor}

\usepackage{booktabs}
\usepackage{subfigure}
\usepackage{tabularx}

\def\BibTeX{{\rm B\kern-.05em{\sc i\kern-.025em b}\kern-.08em
    T\kern-.1667em\lower.7ex\hbox{E}\kern-.125emX}}
\begin{document}

\title{Exploring the Long Short-Term Dependencies to Infer Shot Influence in Badminton Matches}


\author{
\IEEEauthorblockN{Wei-Yao Wang\textsuperscript{\rm 1}, Teng-Fong Chan\textsuperscript{\rm 1}, Hui-Kuo Yang\textsuperscript{\rm 1, 3}, Chih-Chuan Wang\textsuperscript{\rm 1}, Yao-Chung Fan\textsuperscript{\rm 2}, Wen-Chih Peng\textsuperscript{\rm 1}}
\IEEEauthorblockA{
\textsuperscript{\rm 1}National Yang Ming Chiao Tung University, Hsinchu, Taiwan \\
\textsuperscript{\rm 2}National Chung Hsing University, Taichung, Taiwan \\
\{sf1638.cs05, tfchan.cs07g, wangcc, wcpeng\}@nctu.edu.tw, \textsuperscript{\rm 3}hgyang@gmail.com, \textsuperscript{\rm 2}yfan@nchu.edu.tw}






}

\maketitle

\begin{abstract}
Identifying significant shots in a rally is important for evaluating players' performance in badminton matches.
While there are several studies that have quantified player performance in other sports, analyzing badminton data is remained untouched. In this paper, we introduce a badminton language to fully describe the process of the shot and propose a deep learning model composed of a novel short-term extractor and a long-term encoder for capturing a shot-by-shot sequence in a badminton rally by framing the problem as predicting a rally result. Our model incorporates an attention mechanism to enable the transparency of the action sequence to the rally result, which is essential for badminton experts to gain interpretable predictions. Experimental evaluation based on a real-world dataset demonstrates that our proposed model outperforms the strong baselines.
The source code is publicly available at https://github.com/yao0510/Shot-Influence.
\end{abstract}

\begin{IEEEkeywords}
sport analytics, badminton language representation, shot influence, attention mechanism
\end{IEEEkeywords}

\section{Introduction}
\label{sec:introduction}


In recent years, a growing body of research has started to explore applying artificial intelligence to the sports industry due to the availability of data and the advancement of techniques.
Such applications not only play an important role in the moment of matches but also have a great influence on the training stage. Because of the considerable number of professional matches and the enormous number of spectators, sports activities such as soccer attract the most attention.
For example, several studies adopting performance analysis \cite{DBLP:conf/kdd/RuizPWL17}, tactics discovery \cite{DBLP:conf/kdd/DecroosHD18}, actions valuing \cite{DBLP:conf/kdd/DecroosBHD19} and similar play retrieval \cite{DBLP:conf/kdd/WangLCJ19} have been carried out.
Other sports research involving the evaluation of actions in basketball \cite{DBLP:conf/kdd/SiciliaPG19} and pattern recognition of tennis \cite{DBLP:conf/sii/MiyaharaTN19} has also been conducted. 

In the field of badminton, measuring how good the shot in a rally is is important for decision-making and tactic investigation. There is still room for exploration in the task of quantifying player performance. Existing approaches in the literature \cite{gomez2020serving, gomez2020long, Wang2020badminton} focus on notational and temporal variables, utilizing statistical data to investigate performance such as forced or unforced errors for analyzing the behavior of a player and evaluating how likely it is that a player can win.
Moreover, analyzing the shot quality and advanced tactics usually focuses on the last few shots before scoring.

These approaches suffer from four limitations.
First, existing approaches label what they want based on their own objectives. This causes redundant efforts of designing formats if there are new studies from different groups. Second, most approaches fail to account for the details of each shot throughout the whole rally. Third, statistic-based methods ignore contextual information between the shots.
The complexity of calculating each situation will increase considerably due to the number of possible results of each shot.
Besides, assessment of experts' advanced tactics requires repeated review of the broadcast videos to discover the patterns, which is time consuming.
Fourth, objectively measuring the influence of the shot is difficult for non-experts since it requires the ability of both long-term and short-term dependencies between the shots.


For addressing the limitations, we propose an unified language for describing badminton plays, which provides a uniform mechanism for translating match videos into dataset for analysis.
Further, based on this proposed language, we built our dataset by manually labeling players' shot actions from broadcast videos of badminton matches held between 2018 and 2020.
Specifically, the basic instance of our data is one-shot action. By recording the shots made between opposing players from serving to score, the sequence of shots forms a rally. Developing from shot to rally enables the interpretation without reviewing videos, and benefits deep learning algorithms on time-series tasks to bridge the gap between computer science and the badminton field.

In this paper, we propose a badminton rally analysis application by leveraging the deep learning technology, which measures the influences of shots based on the actions sequence from the rally.
By making use of the detailed information of each shot and identifying action-outcome correlations, it helps enhance the confidence of decision-making for badminton players' performance improvement.

In summary, the main contributions of our paper are to:
\begin{enumerate}
  \item Propose a language to represent badminton from shot to rally;
  \item Design and develop a deep learning model which captures both short-term and long-term dependencies in a badminton rally to measure shot influences;
  \item Introduce an attention mechanism into the model to enable the transparency of the model on action sequences with respect to rally outcomes.;
  \item Conduct extensive experiments on a real-world dataset to show our model's capability to infer shot influence.
\end{enumerate}


\section{BLSR: A Language for Representing Badminton from Shot to Rally}
\label{sec:language}
In sports data analysis, data is generally manually labeled by professional experts.
Specifically, through reviewing match videos, professional experts compose the video contents into unified data format. For example, for soccer game analysis, SPADL \cite{DBLP:conf/kdd/DecroosBHD19} is proposed to unify the existing event stream data formats for improving the data analysis performance. There are also companies such as Dartfish\footnote{https://www.dartfish.com/} which provide some techniques to analyze and record data from various sports.

However, to our best knowledge, there is no ready-to-go format designed for badminton analysis. As can be expected, directly using existing format, e.g., SPADL, is not feasible due to different sport natures. Also, vendors design their own format which may vary according to different objectives.




To address the aforementioned issues, by consulting with badminton experts, we propose BLSR (\textbf{B}adminton \textbf{L}anguage from \textbf{S}hot to \textbf{R}ally) as a representation to formalize event stream data. The goal of our language is to be human-readable, general, professional but simple. These perspectives allow badminton players and coaches to easily interpret the process without reviewing match videos.

Specifically, BLSR is a language for describing the process of a rally.
A match is composed of two to three sets, each of which has a number of rallies, and each rally consists of several shots by two players.
This characteristic is similar to the composition of a corpus in natural language. Therefore, inspired by the relation of corpus, sentence, and word in natural language, we treat a shot as a word and a rally as a sentence.
Specifically, a rally $i$ can be represented as a sequence of shots $\{\boldsymbol{s}^{(i)}_1, \boldsymbol{s}^{(i)}_2, \cdots, \boldsymbol{s}^{(i)}_{N^{(i)}}\}$ and a tuple of the rally information $\boldsymbol{R}^{(i)}$, where ${N^{(i)}}$ is the total number of shots in that rally.
Each shot $\boldsymbol{s}^{(i)}_n$ is a tuple of eight attributes described as follows: 1) \textit{Player:} the player who performed the shot; 2) \textit{Timestamp:} the shot's hitting time; 3) \textit{Type:} the type of shot; 4) \textit{Back\_hand:} hit the shuttle with back hand or not; 5) \textit{Around\_head:} hit the shuttle around the head or not; 6) \textit{Hit\_area:} the location where the racket hits the ball; 7) \textit{Player\_area:} the location of the player who performed the shot; 8) \textit{Opponent\_area:} the location of the player who prepared to receive the shot.
Each piece of rally information $\boldsymbol{R}^{(i)}$ can be described as follows: 1) \textit{Roundscore\_A:} Player A's current score in the set; 2) \textit{Roundscore\_B:} Player B's current score in the set; 3) \textit{Getpoint\_player:} the player who won the rally; 4) \textit{End\_reason:} the reason why the rally ended.

To be capable of different usages, we record the timestamp of each shot event, which can not only be used for basic applications such as video replays, but can also be utilized as additional features for data analysis.
To describe the shot types accurately, we consulted with professional coaches and players, and defined the 18 shot types as
\textit{
  net shot; 
  return net; 
  smash; 
  wrist smash; 
  lob; 
  defensive return lob; 
  clear; 
  drive; 
  driven flight; 
  back-court drive; 
  drop; 
  passive drop; 
  push; 
  rush; 
  defensive return drive; 
  cross-court net shot; 
  short service;} 
  and \textit{long service. 
}
These shot types give specific and clear definitions such that similar hits will be in the same class.

\begin{figure}
  \centering
  \includegraphics[width=0.95\linewidth]{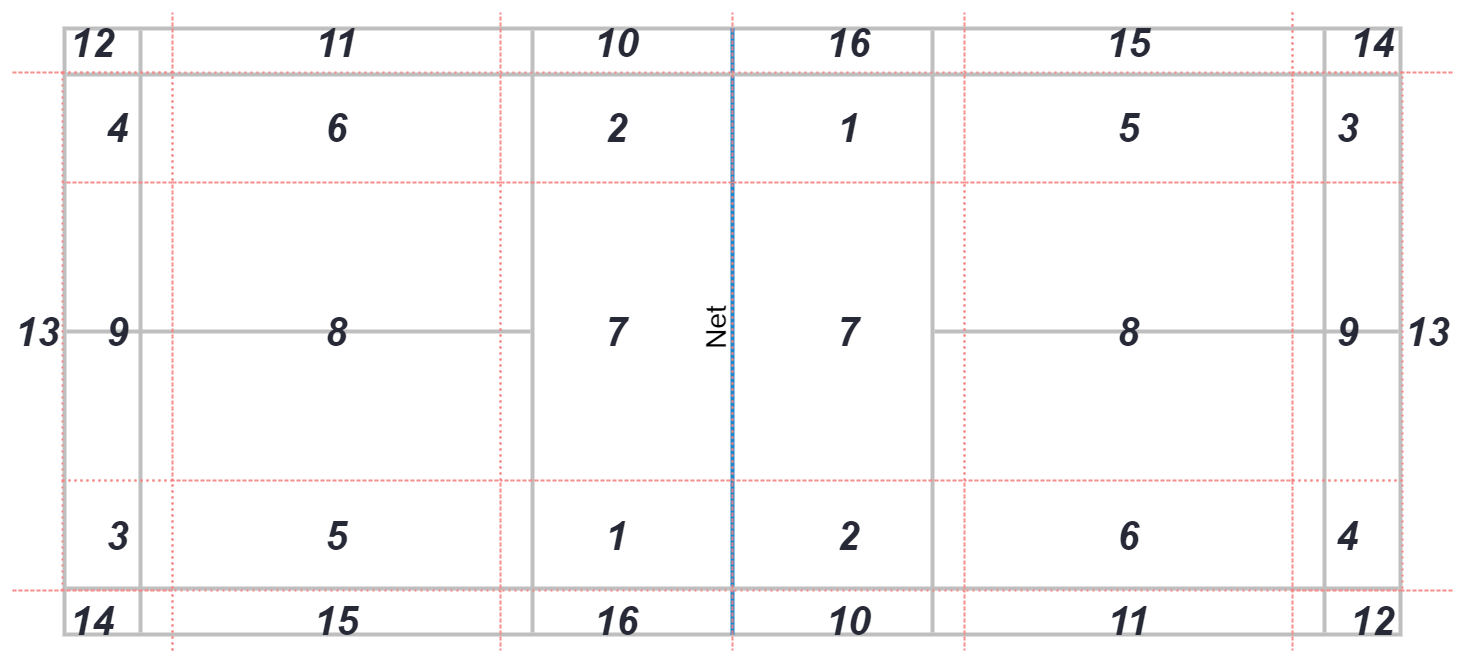}
  \caption{The grid system layout used to represent locations on the badminton court. The blue line represents the net.}
  \label{fig:method-court}
\end{figure}

On the other hand, the location where players perform and receive shots is critical information for detecting tactics.
To alleviate sampling error in the form of coordinates also mentioned in \cite{DBLP:conf/kdd/WangLCJ19}, we discussed with domain experts and proposed a grid system designed for badminton courts to represent the location.
In this grid system, a half-court including the outside part is divided into 16 areas as shown in Fig.~\ref{fig:method-court}.
Note that the grid area is symmetrical with the net in the middle to unify the location information for shots made on different sides.
This grid system is used to represent all of the locations in our data.

Generally speaking, the reason for causing rally end can be classified into five different possible results.
The possible results are \textit{in; out; touch net; not pass over net} and \textit{misjudge}.
The two cases \textit{in} and \textit{out} represent the case that the last shot lands within or outside the court of the opponent's side.
The \textit{touch net} and \textit{not pass over net} are both cases of hitting the shuttle but failing to get it over the net.
Their major difference is whether the shuttle touches the net.

\section{Problem formulation}
\label{sec:problem}

With BLSR to describe a badminton rally, all information can be presented in a structured manner.
Generally speaking, a shot is usually assessed effectively depending on whether the player wins the rally. Therefore, to objectively quantify how good a shot is, this problem can be framed as predicting the final outcome of a rally as follows:

Given a set of rallies \{($\boldsymbol{r}^{(i)}, \boldsymbol{R}'^{(i)}, y^{(i)}$)\}, where $\boldsymbol{r}^{(i)}$ represents a sequence of shots $\{\boldsymbol{s}^{(i)}_1, \boldsymbol{s}^{(i)}_2, \cdots, \boldsymbol{s}^{(i)}_{N^{(i)}}\}$,  $\boldsymbol{R}'^{(i)}$ only contains information not related to the result (with \textit{End\_reason} and \textit{Getpoint\_player} eliminated from $\boldsymbol{R}^{(i)}$, the score only describes the condition at the beginning of the rally), and $y^{(i)}$ indicates the outcome (win or lose) of a player in the rally, our goal is to learn the correlation between player actions, rally state, and outcome. Specifically, we aim to approach a rally's outcome by means of the shots made in that rally and the game progress.

\section{Method}
\label{sec:method}

\begin{figure*}
  \centering
  \includegraphics[width=0.95\linewidth]{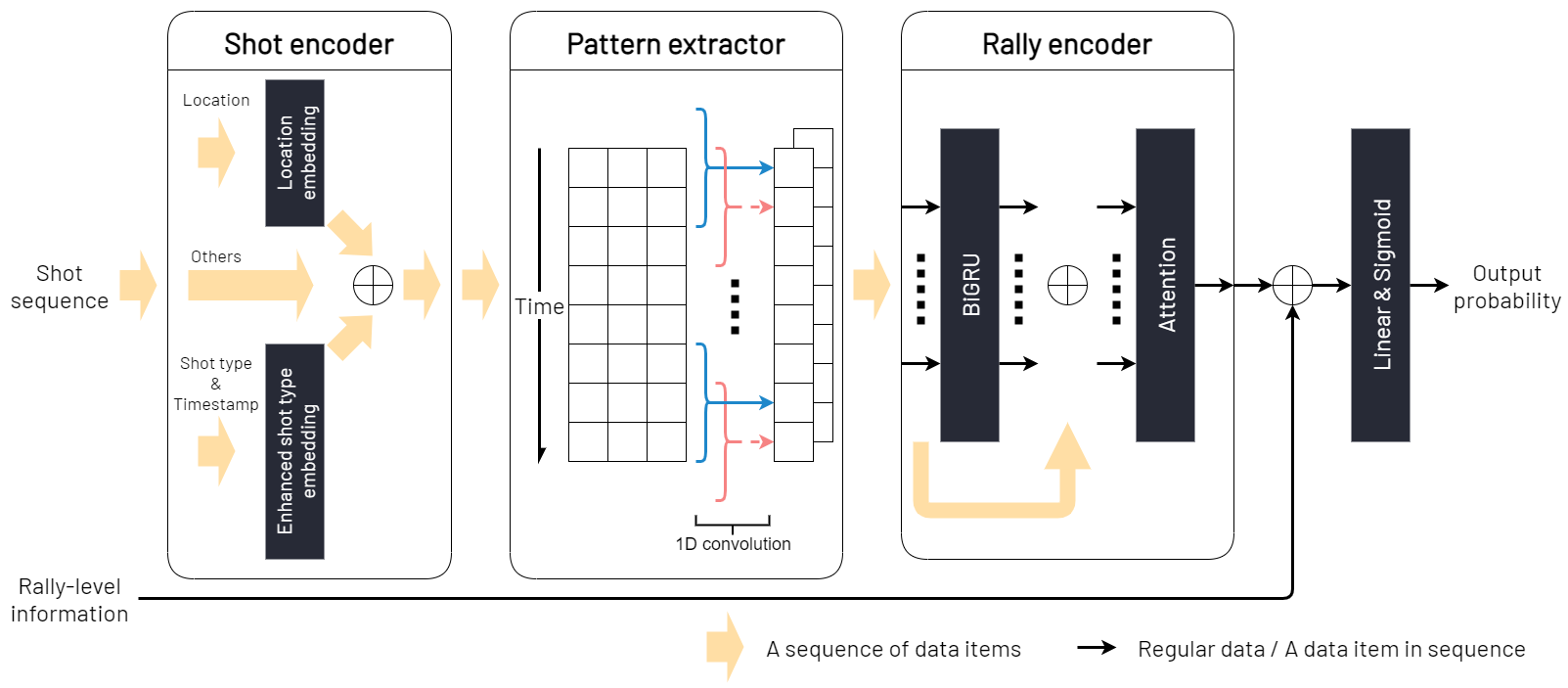}
  \caption{Architecture of the proposed model. The blue solid line and red dashed line in the pattern extractor represent individual convolution operations from two different 1-D convolution neural networks}
  \label{fig:method-framework}
\end{figure*}

Fig.~\ref{fig:method-framework} shows a graphical pipeline of the proposed model.
Our model consists of three major components for predicting the outcome of a rally.
The first component is to encode the shot data (Subsection~\ref{sec:encoding-shot-data}), the second component is to extract short-term pattern information from the processed sequence (Subsection~\ref{sec:extracting-local-shot-patterns}), and the last component is to encode the pattern sequence (Subsection~\ref{sec:encoding-pattern-sequence}).

\subsection{Encoding Shot Data}
\label{sec:encoding-shot-data}

First, we introduce our designs for transforming the input of shot-level information.
Specifically, we apply embedding methods to encode features of each shot in a rally and obtain a processed shot sequence.

\subsubsection{Location embedding}

A naive method to encode location is one-hot encoding.
However, one-hot encoding cannot preserve contextual information between each of them, while using the embedding layer does take context into account.
Thus, each location is assigned a learnable vector and then concatenates with other shot-level information as the output in the first stage, as shown in Fig.~\ref{fig:method-framework}.

\subsubsection{Enhanced shot type embedding}

We follow the idea that we suggest for the location encoding to encode shot type.
That is, we choose to apply an embedding layer to the shot type to capture the underlying information.

Furthermore, inspired by consideration of irregular time gaps from TASA \cite{DBLP:conf/kdd/PavlovskiGSAKGB20}, we adopt temporal score learning to reflect the progress and influence shot as rally progresses.
The timestamp $t^{(i)}_n$ of the $n$-th shot from rally $i$ is converted to time proportion $\tau^{(i)}_n$ with
\begin{equation}
  \tau^{(i)}_n = \frac{t^{(i)}_n - t^{(i)}_1}{t^{(i)}_{N^{(i)}} - t^{(i)}_1}.
\end{equation}
Normalizing it to a range between 0 and 1, it increases linearly and approaches 1 as the shot occurs closer to the last action in the rally.
In other words, it acts as the relative closeness in time of actions to the end of the rally.
On the other hand, the original shot type is converted into two latent variables $\theta^{(i)}_n, \mu^{(i)}_n$ and then combined with the time proportion to form the temporal score $\delta^{(i)}_n$ of the shot using
\begin{equation}
  \delta^{(i)}_n = \sigma(\theta^{(i)}_n + \mu^{(i)}_n \tau^{(i)}_n),
\end{equation}
where $\sigma$ is the sigmoid function.
The two latent variables $\theta^{(i)}_n, \mu^{(i)}_n$ are learnable embeddings during the training process, and their purposes are to determine the effect of the action itself and the effect of action as the time progresses, respectively.
Finally, this obtained temporal score $\delta^{(i)}_n$ will be multiplied to corresponding shot type embedding similar to \cite{DBLP:conf/kdd/PavlovskiGSAKGB20} to preserve these influences and enhance the quality of the shot type embedding.

\subsection{Extracting Local Shot Patterns}
\label{sec:extracting-local-shot-patterns}

After the shot sequence representation, we further incorporate convolutional neural networks (CNNs) for pattern extraction.
Two individual 1-D CNNs are used to separately focus and obtain patterns of different players.
Specifically, they both perform convolutions with $d_{cnn}$ filters of kernel size $K$ from the beginning to the end of the shot sequence representations.
Note that only the $2k$-th output from the first one and the $(2k+1)$-th output from the second one will be extracted and then merged alternately to generate the merged pattern sequence, $\forall k \geq 0$.
That is, the merged pattern sequence $\boldsymbol{p}^{(i)}_{:,j}$ of rally $i$ from the $j$-th filter is obtained by
\begin{equation}
  \begin{split}
    \hat{\boldsymbol{p}}^{(i)}_{:,j} = &ReLU (\boldsymbol{W}^{[C_1]}_j \ast \boldsymbol{r}'^{(i)}+b_j^{[C_1]}),\\
    \bar{\boldsymbol{p}}^{(i)}_{:,j} = &ReLU (\boldsymbol{W}^{[C_2]}_j \ast \boldsymbol{r}'^{(i)}+b_j^{[C_2]}),\\
    \boldsymbol{p}^{(i)}_{:,j} = &AlternateMerge(\hat{\boldsymbol{p}}^{(i)}_{:,j}, \bar{\boldsymbol{p}}^{(i)}_{:,j})
                               = (\hat{\boldsymbol{p}}^{(i)}_{1,j}, \bar{\boldsymbol{p}}^{(i)}_{2,j}, \hat{\boldsymbol{p}}^{(i)}_{3,j}, \cdots),
  \end{split}
  \label{eq:method-convolution}
\end{equation}
where $\ast$ is the convolution operator, $\hat{\boldsymbol{p}}^{(i)}_{:,j}$ and $\bar{\boldsymbol{p}}^{(i)}_{:,j}$ are outputs from different convolutions, $\boldsymbol{W}^{[C_1|C_2]}$ and $b^{[C_1|C_2]}$ are the learnable parameters in the network, and $\boldsymbol{r}'^{(i)}$ is the concatenation of 
the processed shot-level features of the whole rally.
The padding of zero is used in the convolution process to ensure that the resulting pattern has the same length as the input. In other words, a pattern sequence $p^{(i)} \in \mathbb{R}^{N^{(i)} \times d_{cnn}}$ is obtained at this stage.
The choice of kernel size $K$ only affects the number of shots to form one pattern but does not affect the size of the resulting sequence of patterns.

There are two reasons for the need to extract local patterns. First, we treat a shot as a base unit for a badminton rally. However, from the perspective of players, a sequence of consecutive shots is much more important than only one shot. The sub-sequence of a shot sequence usually represents a shot pattern of the player's tactics.
Therefore, when modeling the shot sequence of a rally, we consider the sub-sequence level patterns by employing 1-D CNNs, which is known to be effective in terms of capturing local features along with time domain in time-series tasks.
Second, the shot sequence of a rally is formed by two players making shots alternately.
This motivates us to consider that the patterns made by the two players are different behaviors and mimic the players' thoughts on short-term influence through two 1-D CNNs.

\subsection{Encoding Pattern Sequence}
\label{sec:encoding-pattern-sequence}

The previous stage only captures the short-term pattern in a shot sequence. However, long-term dependency is also critical in a badminton rally for modeling those global playing strategies.
Therefore, we adopt the Gated Recurrent Unit (GRU) \cite{DBLP:conf/emnlp/ChoMGBBSB14} to extract long-term relationships from the shot pattern sequences. 
We employ bidirectional GRU to learn the ability of understanding the long-term influence in a rally.

The state update operations for forward direction involved in the $n$-th pattern are
\begin{equation}
    \boldsymbol{h}^{(i)}_{n,:} = GRU(\boldsymbol{h}^{(i)}_{n-1,:}, \boldsymbol{p}^{(i)}_{n,:};\boldsymbol{W}^{gru}),
  \label{eq:gru}
\end{equation}
where $\boldsymbol{W}^{gru}$ are the GRU parameters.
$\boldsymbol{h}^{(i)}_{n-1,:}$ is the hidden state output of previous patterns.
The initialization of the hidden state in GRU is filled with zeros.
Each of them is a vector of length $d_{gru}$ indicating the number of units used in the GRU network.
Therefore, these operations produce a hidden state output $h^{(i)} \in \mathbb{R}^{N^{(i)} \times d_{gru}}$.

We further incorporate an attention mechanism \cite{DBLP:conf/emnlp/FelboMSRL17} for constructing the final representation of the sequence, which simultaneously enables the shot pattern transparency by obtaining the attention weights as the significance of each of them with respect to the shot sequence representation.
An attention score $e^{(i)}_n$ for the $n$-th step is calculated by the short-term dependency $p^{(i)}_{n,:}$ and long-term dependency $h^{(i)}_{n,:}$ using skip-connections from the corresponding step as
\begin{equation}
  \label{eq:attention_obtain}
  e^{(i)}_n = (\boldsymbol{p}^{(i)}_{n,:} \oplus \boldsymbol{h}^{(i)}_{n,:}) \boldsymbol{W}^{[A]} + b^{[A]},
\end{equation}
where $\oplus$ represents a concatenate operator.
In the learning phase, $\boldsymbol{W}^{[A]}$ and $b^{[A]}$ are parameters to be optimized.
Next, the scores are passed through a softmax function and are normalized to weighted score $\alpha^{(i)}_n$ as follows:
\begin{equation}
  \label{eq:attention_softmax}
  \alpha^{(i)}_n = \frac{e^{(i)}_n}{\sum_{k=1}^{N^{(i)}}e^{(i)}_k}.
\end{equation}
With the weighted score, we obtain the representation $\hat{\boldsymbol{r}}^{(i)} \in \mathbb{R}^{d_{cnn} + d_{gru}}$ of the whole shot sequence as the weighted sum of the short-term patterns and long-term states with
\begin{equation}
  \label{eq:attention_weighted_sum}
  \hat{\boldsymbol{r}}^{(i)} = \sum_{n=1}^{N_i} \alpha^{(i)}_n (\boldsymbol{p}^{(i)}_{n,:} \oplus \boldsymbol{h}^{(i)}_{n,:}).
\end{equation}

\subsection{Predicting Win Probability}

With the feature representation $\hat{\boldsymbol{r}}^{(i)}$ from the previous stages, we further merge this representation and rally-level information $\hat{\boldsymbol{R}}^{(i)} \in \mathbb{R}^{d_{rally}}$ together by concatenating them, forming an overall rally description of $\mathbb{R}^{d_{cnn} + d_{gru} + d_{rally}}$ describing the complete rally.
This $\hat{\boldsymbol{R}}^{(i)}$ differs from the input $\boldsymbol{R}'^{(i)}$ by transforming the features.
It contains the score difference between the two players and how many scores have been consecutively made so that it provides more details about the game state.
The last step is to transform this vector to a win probability $P^{(i)}_{win}$ by a fully-connected layer using the learnable parameters $\boldsymbol{W}^{[L]}$ and $b^{[L]}$ with a sigmoid activation:
\begin{equation}
  P^{(i)}_{win} = \sigma((\hat{\boldsymbol{r}}^{(i)} \oplus \hat{\boldsymbol{R}}^{(i)}) \boldsymbol{W}^{[L]} + b^{[L]}),
  \label{eq:output_prob}
\end{equation}
where the output is a value between 0 and 1 indicating the win probability of a player.
Specifically, 0.5 is the threshold of this value, meaning that the player has a chance to win if it is higher than 0.5, and is likely to lose otherwise.

\subsection{Learning Objectives}

We use the cross-entropy loss
\begin{equation}
  \mathcal{L}_p = \sum_i (y^{(i)}\log P^{(i)}_{win} + (1 - y^{(i)})\log(1 - P^{(i)}_{win}))
  \label{eq:cross_entropy_loss}
\end{equation}
as one of our learning objectives, which is commonly used for classification tasks.

Finally, we define the total loss by combining both cross-entropy loss and regularization loss $\mathcal{L}_r$ as
\begin{equation}
  \mathcal{L} = \mathcal{L}_p + \lambda\mathcal{L}_r,
  \label{eq:total_loss}
\end{equation}
where $\lambda$ is a hyperparameter to adjust the strength of regularization. Adam is adopted to minimize the loss and optimize our model. An early stopping mechanism is also applied to avoid over-fitting.

\section{Experiment}
\label{sec:experiment}

In this section, we evaluate the performance and discuss case studies of our proposed model on a real-world dataset to answer the following research questions: \textbf{Q1}: Does our model outperform other potential baselines?
\textbf{Q2}: Does each component of the model affect the prediction results? 

\subsection{Experiment Setup}

\subsubsection{Dataset}

Data used in the following experiments were collected from real-world badminton matches\footnote{http://bwf.tv}, which are manually labeled by domain experts using a specially designed labeling tool with our BLSR design.
It should be noted that our dataset is private according to internal corporate policies, and the selected players are anonymous to ensure the players' privacy.
Our data contain a total of 15,742 shots, coming from 1,409 rallies which occurred in 19 international men's singles high-ranking matches from 2018 to 2020.
Rallies with missing shot data due to the highlights in the replay video were filtered out from the dataset.
We use one of the players (denoted as Player B, and his opponents, denoted as Player A) in our dataset as the target player to predict his rally performances.
Specifically, his winning rallies are labeled as positive, while the remaining losing rallies are labeled as negative.
We trained our model on 17 out of the 19 matches, approximately 85\% of the total sequences.
The remaining 2 matches were respectively used as a validation set and a testing set with 5\% and 10\% of the total sequences for evaluating the performance of our model.


\subsubsection{Baselines}

To evaluate and compare the performance of our model, we also implemented the following baselines.
We applied the same optimization procedure with default hyperparameter settings on each of them:

\begin{itemize}
  \item Bidirectional LSTM (BiLSTM) is capable of learning long-term dependencies compared with RNN;
  \item Prototype Sequence Network (ProSeNet) \cite{DBLP:conf/kdd/MingXQR19} provides interpretability while keeping the accuracy of the network;
  \item DeepMoji \cite{DBLP:conf/emnlp/FelboMSRL17} is a model designed to understand the sentences and predict the corresponding emotions;
  \item Ordered neurons LSTM (ON-LSTM) \cite{DBLP:conf/iclr/ShenTSC19}.
  It provides a new recurrent architecture that performs neuron ordering to induce latent structures in sequences;
  \item Transformer \cite{DBLP:conf/nips/VaswaniSPUJGKP17}.
  A powerful architecture that is widely attributed to the self-attention mechanism and achieves better performance on long sequence tasks.
\end{itemize}

\subsubsection{Parameter settings}

In the following set of experiments, the dimension of location embedding, shot type embedding, the number of filters in the CNN components, and the number of computational units in the recurrent components were set to 10, 15, 32, and 32, respectively.
All models were trained for 100 epochs.
The default value for CNN kernel size is set to 3, which is suggested by professional badminton experts based on the fact that players and coaches usually treat three shots as a pattern.
For optimization, we initialized the Adam optimizer with a learning rate of 0.001 and set the $\lambda$ regularization hyperparameter to 0.01.
Early stopping was set to monitor the value of AUC.
For each model, we report the average performance on 100 repeated processes.

\subsubsection{Evaluation platform}

All of our evaluation processes were performed on a machine with Intel\textregistered\ Core\texttrademark\ i7-8700 3.2GHz CPU, Nvidia GeForce GTX 2070 GPU, and 32GB RAM, while the methods were implemented in Python~3.6.9 with the Tensorflow~2.0 framework.

\subsection{Prediction Performance}

\subsubsection{Comparing different models}

Since our task was formulated as a classification setting, we aimed to comprehensively evaluate the quality of a predicted probability as an indication of the outcomes.
We used two common evaluation metrics: the brier score (BS) and the area under the receiver operator curve (AUC).
BS is obtained by
\begin{equation}
    BS = \frac{1}{D} \sum_i (P^{(i)}_{win} - y^{(i)})^2,
\end{equation}
where $D$ is the number of sequences in the dataset.

Table~\ref{tab:experiment-prediction-performance-baselines} shows the performance results of our models and the compared baselines.
It can be observed that our model outperformed other baselines on rally outcome prediction in both metrics.
This result suggests that considering dependencies with alternate patterns is more flexible than naive sequence processing to evaluate the performance of the players, and is thus more capable of capturing rich information.
We note that the Transformer model marginally outperformed other baselines, which demonstrated its robust generalization in various domains.
The experiment also shows that using only shot-level data is not sufficient.
It is evident that rally-level data act as extra information for determining the final results, which conforms to the domain's opinions of changing tactics according to different score conditions.

\begin{table}
  \centering
  \caption{Prediction performance compared to different baselines. The best results are in boldface.}
  \label{tab:experiment-prediction-performance-baselines}
  \begin{tabular}{lrr}
    \toprule
    Models                     & AUC             & BS\\
    \midrule
    ProSeNet                   & 0.5333          & 0.2500\\
    DeepMoji                   & 0.7385          & 0.1935\\
    BiLSTM                     & 0.7436          & 0.1859\\
    ON-LSTM                    & 0.7487          & 0.2001\\
    Transformer                & 0.7538          & 0.2009 \\
    \midrule
    Ours w/o Rally-level input & 0.8649          & 0.1476\\
    Ours                       & \textbf{0.8966} & \textbf{0.1329}\\
    \bottomrule
\end{tabular}
\end{table}

\subsubsection{Ablation study}

In order to verify our reasonable design of the proposed model, we conducted an ablation study, and the result is summarized in Table~\ref{tab:experiment-prediction-performance-reduced}.
By experimentally removing specific components from our model, we could obtain an overview of the performance difference and investigate how they affected the model effectiveness.
It is clear that all components had their impact on predicting the outcome of the rally.
Besides, we found that our specially designed CNN components significantly improved the model performance, which provides evidence of showing the significance of mimicking two players' thoughts for short-term influence.
Although the employment of the attention mechanism in the model was mainly designed for providing more interpretable details for the model to decide the final decision, it still slightly improved the performance of both metrics.

\begin{table}
  \centering
  \caption{Prediction performance compared to reduced versions of our model. The best results are in boldface.}
  \label{tab:experiment-prediction-performance-reduced}
  \begin{tabularx}{\linewidth}{Xrr}
    \toprule
    Models                                                & AUC             & BS\\
    \midrule
    Full w/o 2 CNNs                                       & 0.7583          & 0.2117\\
    Full w/o 1 CNN                                        & 0.7596          & 0.2100\\
    Full w/o BiGRU                                        & 0.8903          & 0.1347\\
    Full w/o temporal score enhanced shot type embedding  & 0.8957          & 0.1347\\
    Full w/o Attention                                    & 0.8964          & 0.1339\\
    \midrule
    Full                                                  & \textbf{0.8966} & \textbf{0.1329}\\
    \bottomrule
\end{tabularx}
\end{table}

\section{Related Work}
\label{sec:related-work}



Currently, little research is focusing on badminton data analysis.
Advanced tactics exploration has remained challenging due to the data inaccessibility.
There are several statistic-based approaches \cite{gomez2020serving,gomez2020long} for badminton match analysis which use notational and temporal variables from broadcast videos.
Ghosh et al. \cite{DBLP:conf/wacv/GhoshSJ18} have provided an end-to-end framework for analyzing broadcast badminton match videos.
They detect player actions by using image processing techniques, and provide analytical outcomes based on the identified motions.
Wang et al. \cite{Wang2020badminton} used deep learning techniques to retrieve information from match videos and constructed a platform to visualize their results.
These approaches mainly focus on frequency analysis and fail to illustrate how good a shot is in a rally.
Furthermore, these related works have their own design formats based on their objectives.
Hence, we propose a language to fully describe the rally, and a deep learning model to capture the relationships between the shots.


There are different state-of-the-art approaches which have been performed on sports activities such as soccer and basketball for estimating the outcome of actions.
Decroos et al. \cite{DBLP:conf/kdd/DecroosBHD19} proposed a machine learning model which is applied to soccer event-based data.
It utilizes the action sequence to form game states which are then used to predict the score and concede probabilities.
The difference in these probabilities between game states indicates the value of that action.
Besides, they introduced a language to unify the description of player actions.
Several data science challenges that they posed motivated us to consider designing a strategy.
Sicilia et al. \cite{DBLP:conf/kdd/SiciliaPG19} used a completely different way to achieve a similar purpose for basketball tracking data.
They developed a stacked recurrent model which is used for predicting the probabilities of each outcome from a temporal window of possessions.
The obtained probabilities can then be transformed into an expected score point.
These approaches differ from badminton as a team can perform multiple consecutive actions before any specific outcomes, while the rules of badminton require the player to hit the shuttle and wait for the opponent to return it until it has landed.
While we introduce our model for analyzing badminton data, it should be evident that it can be used for other racket sports (e.g., tennis) as well.



\section{Conclusions and Future Works}
\label{sec:conclusion}

In this paper, we have introduced BLSR, a language to describe the process of the rally.
To measure the influences of the shots, we proposed a deep learning model to capture both long-term and short-term dependencies between those shots.
Also, we introduced an attention mechanism that enables transparency of the model.
The experimental evaluation was conducted on a real-world dataset which shows that our model outperforms other baselines and provides reasonable design from the ablation study.
In our future research, we plan to label more players' data and apply them to our model, and investigate how well other players perform tactics on different opponents.
We aim to use the proposed approach to customize the style and regenerate the players' tactics.

\section*{Acknowledgments}

This work was supported by the Ministry of Science and Technology of Taiwan under Grants MOST-109-2627-H-009-001.

\bibliographystyle{./IEEEtran}
\bibliography{./main}

\end{document}